\documentclass[conference]{IEEEtran}
\IEEEoverridecommandlockouts
\usepackage{cite}
\usepackage{amsmath,amssymb,amsfonts}
\usepackage{algorithmic}
\usepackage{graphicx}
\usepackage{textcomp}
\usepackage{subcaption}
\usepackage{booktabs}
\usepackage{multirow}
\usepackage{url}
\usepackage[table,xcdraw]{xcolor}
\def\BibTeX{{\rm B\kern-.05em{\sc i\kern-.025em b}\kern-.08em
    T\kern-.1667em\lower.7ex\hbox{E}\kern-.125emX}}
\begin{document}

\title{Self-DACE++: Robust Low-Light Enhancement via Efficient Adaptive Curve Estimation
}

\author{\IEEEauthorblockN{1\textsuperscript{st} Jianyu Wen}
\IEEEauthorblockA{\textit{Lenovo Research} \\
Beijing, China \\
jianyuwen2000@gmail.com}
\and
\IEEEauthorblockN{2\textsuperscript{nd} Jun Xie}
\IEEEauthorblockA{\textit{Lenovo Research} \\
Beijing, China \\
xiejun@lenovo.com}
\and
\IEEEauthorblockN{3\textsuperscript{rd} Feng Chen}
\IEEEauthorblockA{\textit{Lenovo Research} \\
Beijing, China \\
chenfeng13@lenovo.com}
\and
\IEEEauthorblockN{4\textsuperscript{th} Zhepeng Wang}
\IEEEauthorblockA{\textit{Lenovo Research} \\
Beijing, China \\
wangzpb@lenovo.com}
\and
\IEEEauthorblockN{5\textsuperscript{th} Chenhao Wu}
\IEEEauthorblockA{\textit{Imperial College London} \\
London, UK \\
cw1323@ic.ac.uk}
\and
\IEEEauthorblockN{6\textsuperscript{th} Tong Zhang}
\IEEEauthorblockA{\textit{Southern University of Science and Technology} \\
Shenzhen, China \\
11911831@mail.sustech.edu.cn}
\and
\IEEEauthorblockN{7\textsuperscript{th} Yixuan Yu}
\IEEEauthorblockA{\textit{Peking University} \\
Beijing, China \\
yuyixuan834@stu.pku.edu.cn}
\and
\IEEEauthorblockN{8\textsuperscript{th} Piotr Świerczyński}
\IEEEauthorblockA{\textit{NODAR Inc.} \\
Somerville, MA, USA \\
pwswierczynski@gmail.com}
}
\maketitle

\begin{abstract}
In this paper, we present \textbf{Self-DACE++}, an improved unsupervised 
and lightweight framework for Low-Light Image Enhancement (LLIE), 
building upon our previous Self-Reference Deep Adaptive Curve Estimation 
(Self-DACE). To better address the trade-off between computational 
efficiency and restoration quality, Self-DACE++ introduces enhanced 
Adaptive Adjustment Curves (AACs). These curves, governed by minimal 
trainable parameters, flexibly adjust the dynamic range while preserving 
the color fidelity, structural integrity, and naturalness of the enhanced 
images. To achieve an extremely lightweight architecture without 
sacrificing performance, we propose a randomized order training strategy 
coupled with a network fusion mechanism, which compresses the model into 
an efficient iterative inference structure. Furthermore, we formulate a 
physics-grounded objective function based on Retinex theory and 
incorporate a dedicated denoising module to effectively estimate and 
suppress latent noise in dark regions. Extensive qualitative and 
quantitative evaluations on multiple real-world benchmark datasets 
demonstrate that Self-DACE++ outperforms existing state-of-the-art 
methods, delivering superior enhancement quality with real-time inference 
capability.
The code is available at \url{https://github.com/John-Wendell/Self-DACE}.
\end{abstract}


\section{Introduction}
\label{sec:intro}

Despite the rapid evolution of imaging sensors, capturing high-fidelity 
images in low-light environments remains a persistent bottleneck for 
computer vision systems. Images taken in suboptimal lighting conditions 
suffer from low visibility, noise corruption, and color distortion, which 
severely degrade the performance of downstream tasks. While recent years 
have witnessed a surge in deep learning-based solutions, achieving a 
balance between restoration quality, real-time processing, and 
cross-domain generalization remains a formidable challenge, especially 
for resource-constrained edge devices.

Existing approaches generally fall into two categories: 
adjustment-curve-based and generative-based methods. While curve-based 
approaches prioritize efficiency, they often struggle with complex, 
non-uniform illumination. Conversely, generative methods---ranging from 
CNN-based networks to recent Transformer and diffusion models---achieve 
higher visual quality but incur substantial computational costs and 
suffer from poor generalization due to their reliance on paired 
supervision.

Building upon our previous work Self-DACE~\cite{wen2023self}, which 
introduced a self-reference unsupervised framework for low-light 
enhancement, we present \textbf{Self-DACE++}, a further improved 
framework that addresses the remaining limitations in restoration 
quality, noise suppression, and generalization. Specifically, while 
Self-DACE demonstrated promising results in lightweight curve-based 
enhancement, it still faces challenges in handling severe noise in 
extremely dark regions and achieving consistent performance across 
diverse real-world scenarios. Self-DACE++ tackles these issues through 
enhanced Adaptive Adjustment Curves (AACs), a more robust physics-grounded 
loss formulation, and an improved denoising module, collectively 
delivering superior enhancement quality while preserving real-time 
inference capability.

Our main contributions are as follows:
\begin{itemize}
    \item We design enhanced AACs, flexible and differentiable functions 
    capable of dynamically adjusting the dynamic range without introducing 
    artifacts or overexposure, offering improved smoothness and fidelity 
    over the original Self-DACE~\cite{wen2023self}.

    \item We propose an improved lightweight framework featuring a 
    randomized order training strategy and an RNN-like iterative inference 
    mechanism. This design significantly reduces the parameter count, 
    making the model well-suited for real-time applications on 
    heterogeneous and resource-constrained devices.

    \item We formulate a refined set of unsupervised loss functions 
    rooted in Retinex theory and the physical camera model, which more 
    effectively constrains the learning process and substantially enhances 
    the model's generalizability across diverse real-world scenarios.

    \item We incorporate a dedicated denoising module that explicitly 
    estimates and suppresses latent noise in dark regions, addressing a 
    key limitation of the original Self-DACE and improving robustness 
    under extremely low-light conditions.

    \item We conduct extensive qualitative and quantitative evaluations, 
    demonstrating that Self-DACE++ surpasses existing state-of-the-art 
    approaches---including the original Self-DACE---in terms of 
    enhancement quality, inference speed, and cross-domain generalization, 
    particularly on unseen real-world datasets.
\end{itemize}

\section{Related Work}
\label{sec:rework}
In this section, we briefly review the two dominant paradigms in LLIE: adjustment-curve-based and generative-based methods.

\noindent{\bf Adjustment-curve-based methods.}
To avoid the heavy computational burden of pixel-wise reconstruction, adjustment-curve-based methods estimate global or local curves to map low-light pixels to a normal range\cite{11227405,10650454,wang2022image,zhang2021rellie,gao2023bezierce,veluchamy2021optimized,yang2023learning,guo2020zero,li2021learning}.
Early works utilized wavelet decomposition~\cite{loza2013automatic} or S-curve adjustments like ExCNet~\cite{zhang2019zero} to handle uneven illumination.
A paradigm shift occurred with ZeroDCE~\cite{guo2020zero}, which formulated enhancement as an image-specific curve estimation problem using quadratic polynomials. While highly efficient, the simplistic curve formulation in ZeroDCE often limits its capacity to handle complex degradation.
Subsequent research has sought to refine this paradigm by  incorporating semantic guidance~\cite{zhang2021rellie}, or employing more complex curve functions such as Bezier curves~\cite{gao2023bezierce,veluchamy2021optimized} and the Naka-Rushton function~\cite{yang2023learning}.
Despite these improvements, balancing curve flexibility with spatial consistency remains a challenge.

\noindent{\bf Generative-based methods.}
Generative methods treat LLIE as a signal reconstruction or generation task, often leveraging Retinex theory to decompose images into reflectance and illumination components.
Pioneering works like LIME~\cite{guo2016lime} introduced the Dark Channel Prior for illumination estimation, inspiring deep learning successors such as KinD~\cite{zhang2019kindling}.
To improve efficiency, architecture search and unfolding techniques were introduced in RUAS~\cite{liu2021retinex} and SCI~\cite{ma2022toward}.
However, the pursuit of higher visual quality has led to increasingly complex architectures.
Structure-revealing methods~\cite{xu2022structure,xu2023low} utilize gradient priors to suppress noise, while others integrate semantic segmentation features~\cite{liang2022semantically,wu2023learning} to guide the enhancement.
Furthermore, Generative Adversarial Networks (GANs)~\cite{jiang2021enlightengan,fu2022gan} and Diffusion models~\cite{hou2024global,Li2025TSDiffTD} have been employed to synthesize realistic textures.
Most recently, Vision Transformers have gained significant prominence\cite{10651180,wang2023ultra,Cai_2023_ICCV,10191264,10650029}. Methods like LLFormer~\cite{wang2023ultra} and Retinexformer~\cite{Cai_2023_ICCV} adapt the self-attention mechanism to capture long-range dependencies.
Although these large-scale models achieve impressive metrics, their heavy computational load and reliance on paired synthetic data often hinder their deployment in real-world, real-time scenarios.

In summary, existing methods face a trade-off between efficiency and quality: curve-based approaches are fast but limited by simplistic formulations, while generative models achieve superior results at the cost of heavy computation and poor generalization.
To address this dilemma, we propose a lightweight, unsupervised framework that integrates flexible Adaptive Adjustment Curves with Retinex-based constraints, achieving real-time performance, robust generalization, and high-fidelity enhancement.
\begin{figure*}[tb]
  \centering
  
  \includegraphics[width=1\linewidth]{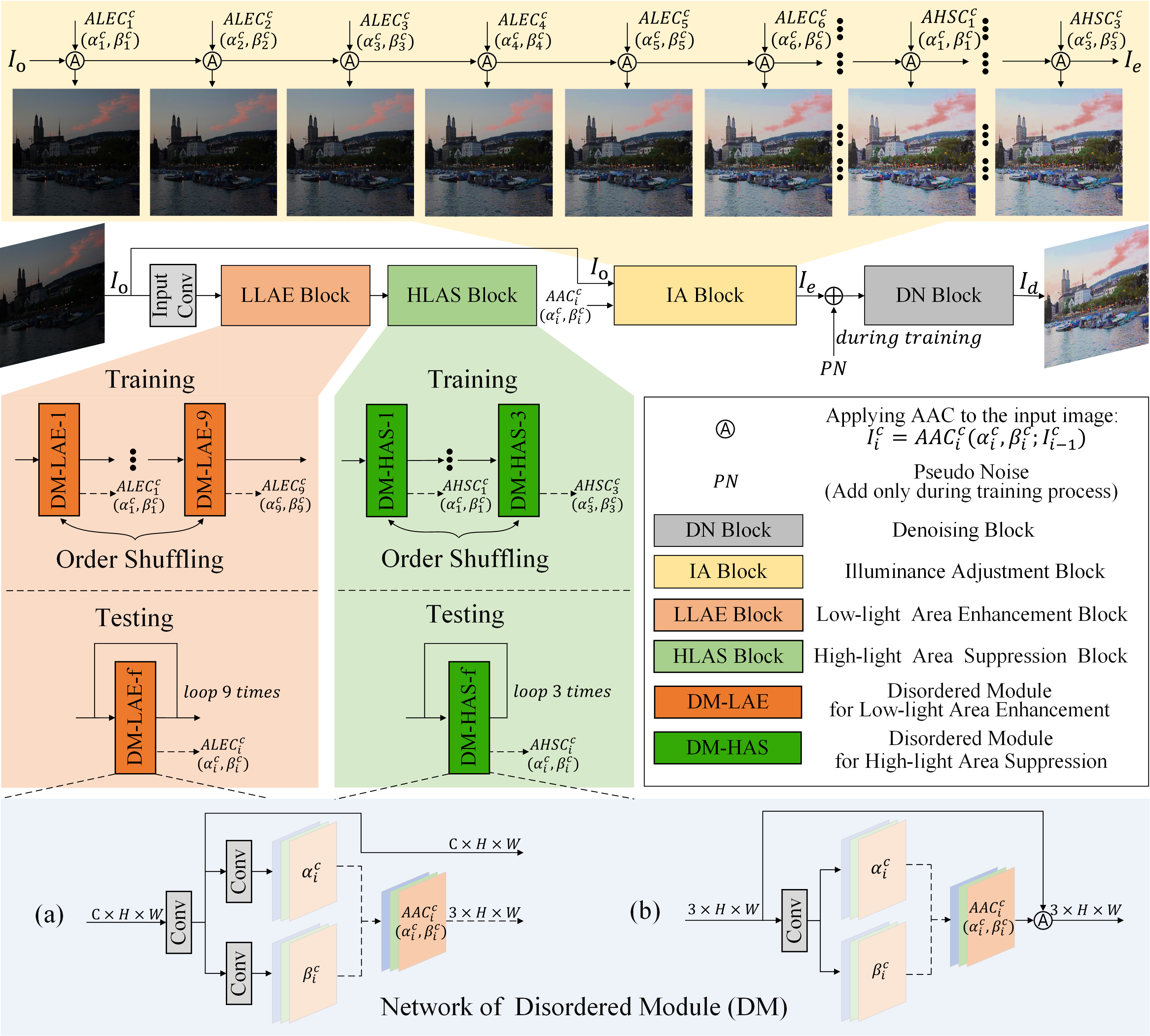} 
\caption{
  The framework of Self-DACE++. 
  (a) The standard DM module employed by both the Standard and Small versions. 
  (b) The lightweight and fast DM module designed specifically for the Tiny version.
}
  \label{fig:frame}  
\end{figure*} 

\section{Methodology}
In this section, we introduce Adaptive Adjustment Curves (AACs) applied iteratively to each image pixel. 
We design unsupervised loss functions based on Retinex theory.
We propose a light framework for dark illuminance adjustment and latent noise removal, outlined in Fig.~\ref{fig:frame}.

\subsection{Adaptive Adjustment Curves}
We aim to design a flexible adjustment curve, which can handle images in complex lighting without artifacts and overexposure. We propose a monotonous, differentiable Adaptive Adjustment Curve (AAC) function, expressed as
\begin{equation}
  AAC{^c}(\alpha{^c},\beta^c; I^c) =  I^c + \alpha^c \otimes 1/\beta^c \otimes C(\beta^c;I^c), 
  \setlength\belowdisplayskip{2pt}
\end{equation}
where, for LAEC,
\begin{equation}
C(\beta{^c};I^c) = S(-k\cdot(I^c-\beta{^c}+\delta)) \otimes I^c\otimes(\beta^c - I^c),
  \setlength\belowdisplayskip{2pt}
\end{equation}
and, for HASC,
\begin{equation}
C(\beta{^c};I^c) = S(k\cdot(I^c-\beta{^c}-\delta)) \otimes (1-I^c)\otimes(I^c - \beta^c).
  \setlength\belowdisplayskip{2pt}
\label{eq:Sig}
\end{equation}
$I^c$ denotes the per-pixel intensity value of $c$-channel normalized to $[0, 1]$, $c$ denotes color channels in the RGB color space, $S$ is the sigmoid function, and $\otimes$ represents element-wise multiplication.
Parameter $\alpha$ adjusts the magnitude of AACs, whereas $\beta$ controls the range of enhancement or suppression. 
For LAECs, $\alpha(\mathbf{x}) \in [0, 1], \beta(\mathbf{x})\in[0.3, 1]$, and for HASCs, $\alpha(\mathbf{x}) \in [-1, 0], \beta(\mathbf{x})\in[0.7, 0.9]$, which all are trainable pixel-wise scalar maps.
$S(\beta;I_c)$ is introduced to suppress parts of $I^c$ with intensity exceeding $\beta$.
What's more, $k$ and $\delta$ can help $S(\beta;I_c)$ suppress AAC in advance to avoid curve oscillation, and we set $k$ as 15 and $\delta$ as 0.1.




\subsection{Framework}

The proposed Self-DACE++ framework, illustrated in Fig.~\ref{fig:frame}, is designed to adaptively adjust image brightness while effectively managing noise. It primarily consists of an Illuminance Adjustment (IA) block followed by a Denoising (DN) block.

\noindent \textbf{Illuminance Adjustment Strategy.} 
The IA block integrates Low-light Area Enhancement (LLAE) and High-light Area Suppression (HLAS) modules to address under- and over-exposure, respectively. To achieve a lightweight design without compromising performance, we introduce a Disordered Module (DM) governed by a specific training-inference paradigm. During the training phase, we employ a randomized order strategy where LLAE and HLAS utilize 9 (DM-LAE) and 3 (DM-HAS) distinct DMs, respectively. By randomizing the sequence, each module is forced to capture both shallow and deep features independently, ensuring consistent convergence. Subsequently, to reduce memory overhead during inference, we fuse the trained modules into a single representative unit (DM-f) via parameter averaging:
\begin{equation}
\left\{
\begin{aligned}
  W_{\text{f}} = 1/K\textstyle\sum_{i=1}^K W_{i}\\
  B_{\text{f}} = 1/K\textstyle\sum_{i=1}^K B_{i},\\
\end{aligned}
\right.
  \label{eq:weight}
  \setlength\belowdisplayskip{2pt}
\end{equation}
where $W$ and $B$ denote weights and biases, and $K$ is the number of modules (9 for LLAE, 3 for HLAS). This fused module operates iteratively in a self-looping manner, effectively compressing a deep network into a recurrent shallow network (RNN-like) for efficient deployment.

\noindent \textbf{Denoising and Noise Simulation.}
Since noise in low-light images is often amplified during enhancement, a DN block is appended after the IA block. To train this block, we simulate Pseudo Noise (PN) in dark regions:
\begin{equation}
  PN = (1 - I^c) \times \mathcal{N}(0,\sigma^c),
  \label{eq:noise}
  \setlength\belowdisplayskip{2pt}
\end{equation}
where $\mathcal{N}(0,\sigma^c)$ represents a Gaussian distribution. During training, $\sigma$ is randomly sampled from $[1, 5]$.

\noindent \textbf{Model Variants.}
To accommodate diverse computational constraints, we provide three variants:
\begin{itemize}
    \item \textbf{Standard Version:} Utilizes the full DM (Fig.~\ref{fig:frame}(a)) with the DN block for high-fidelity enhancement.
    \item \textbf{Small Version:} Adopts the standard DM but removes the DN block, prioritizing real-time efficiency for scenarios with less severe noise.
    \item \textbf{Tiny Version:} Employs a simplified DM (Fig.~\ref{fig:frame}(b)) to minimize model size for ultra-lightweight deployment.
\end{itemize}

\subsection{Loss Function Design}
Drawing on Retinex theory, we take a normalization approach accounting for the intensity of three color channels, and define the illuminance $L$ and reflectance $R$  as
\begin{equation}
\left\{
\begin{aligned}
L &= I^r + I^g + I^b\\
R^c &= I^c \oslash (L + \varepsilon), \quad \varepsilon > 0,\\
\end{aligned}
\right.
\setlength\belowdisplayskip{2pt}
\end{equation}
where $\oslash$ stands for pixel-wise division. We add a small offset $\varepsilon = 10^{-4}$ to avoid division by $0$.

\noindent {\bf Reflectance Consistency Loss.} The first component of the loss function stems from an assumption based on Retinex theory that the reflectance map is invariant to illuminance. 
We thus enforce the reflectance map of an enhanced image to be close to that of the original image. We propose $L_{RC}$ to preserve the color distribution during the enhancement process:
\begin{equation}
\textstyle
  L_{RC} = \sum_{c \in \lbrace r, g, b \rbrace}\Vert R{^c_o} - R{^c_e} \Vert_2^2,
  \label{eq:lcol}
   \setlength\belowdisplayskip{2pt}
\end{equation}
where $R{^c_o}$ and $R{^c_e}$ denote the pixel-wise reflectance of the low-light image $I_o$ and enhanced image $I_e$, respectively.

\noindent {\bf White Balance Loss.} To avoid over-saturation which most methods suffers, we develop $L_{WB}$ to control the white balance. We design $L_{WB}$ as
\begin{equation}
\textstyle
  L_{WB} = \sum_{c \in \lbrace r, g, b \rbrace} \Big( A{^c_e}-\frac{1}{3} \Big)^2.
  \label{eq:gcol}
    \setlength\belowdisplayskip{2pt}
\end{equation}
with
\begin{equation}
\textstyle
    A{^c} = \Big(\sum_{n=1}^N I^c_{(n)}\Big) / \Big(\sum_{n=1}^N \sum_{c' \in \lbrace r, g, b \rbrace} I^{c'}_{(n)}\Big) .
    \setlength\belowdisplayskip{2pt}
\end{equation}
Here, $A{^c}$ is the average value of intensity for $c\in \lbrace r, g, b \rbrace$ channel in the whole image.
And, $N$ is the number of pixels in an image. $A{^c_e}$ is $A{^c}$ for the enhanced image.

\begin{table*}[tb]
\caption{Quantitative comparisons in terms of $\mathcal{P}$ (PSNR), $\mathcal{S}$ (SSIM), $\mathcal{L}$ (LPIPS), $\bigtriangleup$ (CIEDE2000), and Average Precision (AP).
We evaluate the face detection performance on the DarkFace dataset enhanced by different methods using RetinaFace with IoU thresholds of 0.25, 0.50, and 0.75.
The table is organized into four sections: Supervised methods, followed by Unsupervised methods categorized into Large, Small, and Tiny scales based on model complexity and inference speed.
\textbf{The best} result is in bold, \underline{the sub-optimal} result is underlined. 
$\circ$ denotes the results are trained on LOL and tested on SCIE dataset.
Results marked with $\bullet$ are retrained and tested on SCIE dataset by us.
}

\resizebox{\linewidth}{!}{
\label{tab:comparison}
\centering
\begin{tabular}{c|ccc|cccc|cccc|ccc}
\toprule
\makebox[0.01\textwidth][c]{\multirow{2}{*}{Methods}}   
& \multirow{2}{*}{\begin{tabular}[c]{@{}c@{}}Params\\ 
(M)↓\end{tabular}} 
& \multicolumn{1}{c}{\multirow{2}{*}{\begin{tabular}[c]{@{}c@{}}FLOPs\\ 
(G)↓\end{tabular}}}   
& \multicolumn{1}{c|}{\multirow{2}{*}{\begin{tabular}[c]{@{}c@{}}FPS\\ 
(F/S)↑\end{tabular}}}   
& \multicolumn{4}{c|}{LOL-test}                                               
& \multicolumn{4}{c|}{SCIE-part2} 
& \multicolumn{3}{c}{DarkFace}
\\
& 
&   
&        
& $\mathcal{P}$↑               & $\mathcal{S}$↑             & $\mathcal{L}$↓               & $\bigtriangleup$↓             
& $\mathcal{P}$↑               & $\mathcal{S}$↑             & $\mathcal{L}$↓               & $\bigtriangleup$↓            
& AP(0.25)↑               & AP(0.50)↑             & AP(0.75)↑              
\\
\midrule\multicolumn{15}{c}{Supervised Methods} \\ \midrule
KinD++\cite{zhang2021beyond}       & \underline{1.068}      & 12238.03 & 0.46       & 17.75            & 0.77          & 0.20            & 14.74          & 20.06                                                                 & 0.75                                                                 & \underline{0.43}                                                           & 11.63        & 0.459 & 0.295 & 0.007
\\
URN\cite{wu2022uretinex}          & \textbf{0.340}    & 938.23   & \textbf{17.14}
& 19.86            & 0.83          & \textbf{0.13}   & 12.39        & $\circ$21.47                                                                 & $\circ$0.77                                                                 & $\circ$0.44                                                                 & $\circ$9.83     & \underline{0.533} & \underline{0.351} &   \textbf{0.182}                                                           \\
SNR-Net\cite{xu2022snr}      & 39.124           & \underline{394.10}  & 1.76   & \underline{24.61}      & \underline{0.84}    &  \underline{0.15}            & \underline{7.17}     & \begin{tabular}[c]{@{}c@{}}$\circ$15.48\\ $\bullet$\underline{22.15}\end{tabular} & \begin{tabular}[c]{@{}c@{}}$\circ$0.65\\ $\bullet$\underline{0.80}\end{tabular} & \begin{tabular}[c]{@{}c@{}}$\circ$0.56\\ $\bullet$0.45\end{tabular} & \begin{tabular}[c]{@{}c@{}}$\circ$24.55\\ $\bullet$\underline{9.36}\end{tabular}  & 0.265  & 0.182  & 0.004\\
Retinexformer\cite{Cai_2023_ICCV}& 1.606            & \textbf{280.45}& \underline{3.22} & \textbf{25.15}   & \textbf{0.85} & \textbf{0.13}      & \textbf{6.34}  & \begin{tabular}[c]{@{}c@{}}$\circ$17.22\\ $\bullet$\textbf{23.22}\end{tabular} & \begin{tabular}[c]{@{}c@{}}$\circ$0.74\\ $\bullet$\textbf{0.81}\end{tabular} & \begin{tabular}[c]{@{}c@{}}$\circ$0.45\\ $\bullet$\textbf{0.40}\end{tabular} & \begin{tabular}[c]{@{}c@{}}$\circ$18.11\\ $\bullet$\textbf{8.05}\end{tabular}   & \textbf{0.576}  & \textbf{0.356}  & \underline{0.008} \\
\midrule \multicolumn{15}{c}{Unsupervised Methods (Large)}\\ \midrule


EnGAN\cite{jiang2021enlightengan}& 8.637            & \textbf{273.24}    & 17.96      & 17.48            & 0.65          & 0.32            & 17.92          & 17.81                                                                 & 0.74                                                                 &  \textbf{0.43}                                                                 & 16.54  & \textbf{0.538}  & \underline{0.344}  & \textbf{0.008}                                                              \\
PairLIE\cite{Fu_2023_CVPR}   & \underline{0.342}            & \underline{368.27}     & \textbf{31.16}     & \underline{19.51}      & \underline{0.74}    & \underline{0.25}      & \textbf{12.60} & \underline{20.31}                                                                 & \textbf{0.79}                                                        & \underline{0.45}                                                                 & \underline{11.23}    & \underline{0.519}  & \textbf{0.388}  & 0.006                                                           \\
\rowcolor[HTML]{C0C0C0}
 \textbf{Ours}                                                                   & \textbf{0.654}         & 764.23     & \underline{19.44}   & \textbf{19.69} & \textbf{0.78} & \textbf{0.18} & \underline{13.80} & \textbf{21.02}                                                           & \underline{0.75}                                                           & 0.46                                                                 & \textbf{10.73}        & 0.508  & 0.334  & \underline{0.007}                                              \\
\midrule\multicolumn{15}{c}{Unsupervised Methods (Small)} \\ \midrule
 RRDNet\cite{zhu2020zero}     &   0.128            & 137.95      & 0.45     & 10.92            & 0.43          & 0.38            & 49.47          & \underline{14.98}                                                                 & 0.67                                                                 & 0.55                                                                 & \underline{22.61}                                & 0.571  & 0.311  & 0.004                                 \\
ZeroDCE\cite{guo2020zero}& \underline{0.079}            & \underline{84.99}     & \textbf{77.74}      & \underline{14.86}            & \underline{0.56}          & \underline{0.34}            & \underline{25.47}          & 14.81                                                                 & \underline{0.69}                                                                 & \underline{0.44}                                                                 & 24.16                                                  & \underline{0.655}  & \underline{0.382}  & \underline{0.006}                \\
\rowcolor[HTML]{C0C0C0}
  \textbf{Ours-Small}                                                &\textbf{0.023}          & \textbf{83.67} & \underline{35.66}          & \textbf{18.91}            & \textbf{0.59}          & \textbf{0.33}            & \textbf{14.37}          & \textbf{21.03}                                                        & \textbf{0.75}                                                                 & \textbf{0.43}                                                        & \textbf{10.78}                                         & \textbf{0.666}  & \textbf{0.395}  & \textbf{0.009}                  \\
\midrule \multicolumn{15}{c}{Unsupervised Methods (Tiny)} \\ \midrule
RUAS\cite{liu2021retinex}     & 0.003            & 3.53    & 11.63        & \underline{16.40}            & 0.50          & \textbf{0.27}            & \underline{22.90}          & \underline{14.98}                                                                 & \underline{0.67}                                                                 & 0.55                                                                 & \underline{22.61}                                                       & \underline{0.638}  & \underline{0.372}  & \underline{0.006}          \\
SCI\cite{ma2022toward}          & \underline{0.00035}    & \textbf{0.58}  & \underline{26.60} & 14.78            & \underline{0.52}          & 0.34            & 30.41          & 14.07                                                                 & 0.65                                                                 & \textbf{0.43}                                                        & 29.13                                               & 0.628  & 0.358  & 0.005                  \\

\rowcolor[HTML]{C0C0C0}
  \textbf{Ours-Tiny}                                             & \textbf{0.00034} & \underline{2.10}     & \textbf{51.37}  & \textbf{17.65}            & \textbf{0.61}          & \underline{0.32}            & \textbf{16.25}          & \textbf{19.85}                                                                 & \textbf{0.74}                                                                 & \underline{0.45}                                                                 & \textbf{12.20}   & \textbf{0.663}  & \textbf{0.393}  & \textbf{0.007} \\
\bottomrule
\end{tabular}
}
\end{table*}

\begin{table}[tb]
    \centering
    \setlength{\tabcolsep}{1.5pt} 
    \begin{minipage}[t]{0.48\linewidth} 
        \centering
        \caption{Ablation study of HLAS and LLAE block.}
        \label{tab:abl-llae-hlas}
        \resizebox{\linewidth}{!}{
            \begin{tabular}{cccc}
                \toprule 
                Block & \#ch - \#iter & PSNR↑ & SSIM↑ \\ 
                \midrule
                \multirow{5}{*}{LLAE} & 32 - 8  & 19.61 & 0.75  \\ 
                                      & 16 - 9  & 19.43 & 0.70  \\ 
                                      & 32 - 9  & 19.69 & 0.78  \\ 
                                      & 64 - 9  & 19.70 & 0.78  \\ 
                                      & 32 - 10 & 19.70 & 0.78  \\ 
                \midrule
                \multirow{6}{*}{HLAS} & 32 - 2  & 19.62 & 0.74  \\ 
                                      & 16 - 3  & 19.65 & 0.75  \\ 
                                      & 32 - 3  & 19.69 & 0.78  \\ 
                                      & 64 - 3  & 19.69 & 0.78  \\ 
                                      & 32 - 3  & 19.64 & 0.77  \\  
                                      & w/o HLAS& 19.47 & 0.73  \\ 
                \bottomrule
            \end{tabular}
        }
    \end{minipage}
    \hfill 
    \begin{minipage}[t]{0.5\linewidth} 
        \centering
        \caption{Ablation study of each loss component.}
        \label{tab:loss_ab}
\renewcommand{\arraystretch}{1.6} 
        \resizebox{\linewidth}{!}{
            \begin{tabular}{ccc} 
                \toprule
                Ablation term & PSNR↑  & SSIM↑  \\ 
                \midrule
                w/o $L_{IL}$  & 7.91  & 0.40  \\ 
                w/o $L_{RC}$ & 10.21 & 0.57  \\ 
                w/o $L_{WB}$ & 19.51 & 0.74  \\ 
                w/o $L_{CS}$ & 14.21 & 0.50  \\ 
                \midrule
                w/o $\mathcal{S}$  & 19.21 & 0.74  \\ 
                w/o $\Vert \bigtriangledown I_e - \bigtriangledown I_d\Vert_2^2$ & 19.47 & 0.75  \\ 
                w/o $\Vert \bigtriangledown I_d\Vert_2^2$ & 19.56 & 0.77  \\ 
                \bottomrule
            \end{tabular}
        }
    \end{minipage}
\end{table}

\noindent {\bf Illuminance Loss.} 
Based on the physics laws of illumination, white objects could reflect more light than other colored objects. Therefore, we design a pixel-wise illuminance estimator $E$  to estimate the illumination with reference to the distance to white color distribution, $[\frac{1}{3},\frac{1}{3},\frac{1}{3}]$.
$E$ is defined as
\begin{equation}
\textstyle
  E = 1-\sum_{c \in \lbrace r, g, b \rbrace } \big\Vert R{^c_o} - \frac{1}{3} \big\Vert_2.
  \label{eq:H}
  \setlength\belowdisplayskip{2pt}
\end{equation}
Then, the illuminance loss $L_{lum}$ is designed as
\begin{equation}
  L_{IL} = \big\Vert y\cdot E - L_e\big\Vert_2^2,
  \setlength\belowdisplayskip{2pt}
  \label{eq:lum}
\end{equation}
where $L_e$ represents the illuminance map of the output enhanced image $I_e$,
and $y$ is the expected illuminance level, which is set manually set on demand.
We choose $y = 0.8$.

\noindent {\bf Curve Smoothness Loss.} To preserve the spatial structure and consistency of the image and avoid undesirable spatial artifacts, we propose a loss function enforcing the adjustment curve smoothness in the spatial dimension. More precisely, we want to minimize the total variation of the $\alpha$ and $\beta$ used in the definition of AAC. This loss is defined as follows
\begin{equation}
\textstyle
  L_{CS}(\zeta)=1/N\sum_{c\in\lbrace r, g, b \rbrace}\Vert\bigtriangledown\zeta^c\Vert_2^2, \zeta\in\{\alpha,\beta\}.
  \setlength\belowdisplayskip{2pt}
  \label{eq:smooth}
\end{equation}

\noindent {\bf Denoising Loss Function.} Denoising block aims to estimate enhanced noise previously hidden in the dark images. 
$I_d$ is the enhanced result after denoising a simulated noisy image, and $I_e$ is obtained by enhancing the noise-free image. Denote SSIM function as $\mathcal{S}$.
To this end, we design the loss as 
\begin{equation}
  L_{DN} = -w_s\mathcal{S}(I_e, I_d) + w_g\Vert \bigtriangledown I_e - \bigtriangledown I_d\Vert_2^2 + \Vert \bigtriangledown I_d\Vert_2^2,
   \setlength\belowdisplayskip{2pt}
  \label{eq:total_2}
\end{equation}

\noindent {\bf Total Loss Function.} The total loss function combines together all the sub-loss functions defined above is
\begin{equation}
  L = w_{R}L_{RC} + w_{W}L_{WB} + w_{I}L_{IL}+ w_{\zeta}L_{CS} +  L_{DN}.
  \setlength\belowdisplayskip{2pt}
  \label{eq:total}
\end{equation}
We set the weights corresponding to different components of the loss function as $w_{R}=20000$, $w_{W}=5$, $w_{I}=10$, $w_{\alpha}=20000$,  $w_s = 10$ and $w_g = 40$ and $w_{\beta}=20000$.

\section{Experiments and Analysis}

In this section, we present implementation details and an ablation study investigating the significance of the main components of our algorithm. We also show quantitative and qualitative comparison between images enhanced using our method and other existing algorithms. 

\subsection{Implementation Details}

We train Self-DACE++ on SCIE~Part1~\cite{cai2018learning} dataset. We resize all training images to $256\times 256$.
LLAE and HLAS are lightweight and trained jointly.
And we trained denoising network independently for 200 epochs.

\begin{figure*}[tb]
\centering
\captionsetup{skip=2pt} 
\captionsetup[subfigure]{skip=2pt} 

\subcaptionbox{Input\label{fig:comparison_lol_input}}{
  \includegraphics[width=0.15\linewidth]{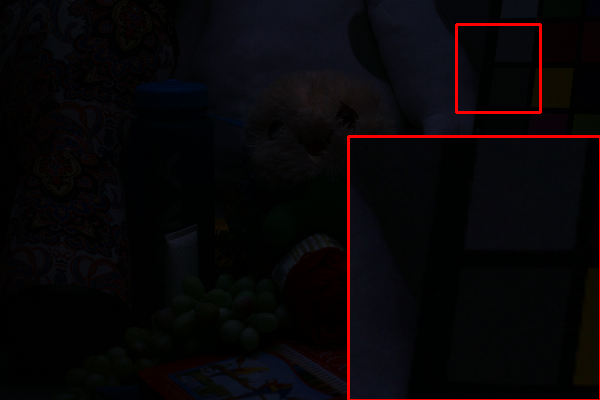}
}\hfill
\subcaptionbox{KinD++\label{fig:comparison_lol_kindp}}{
  \includegraphics[width=0.15\linewidth]{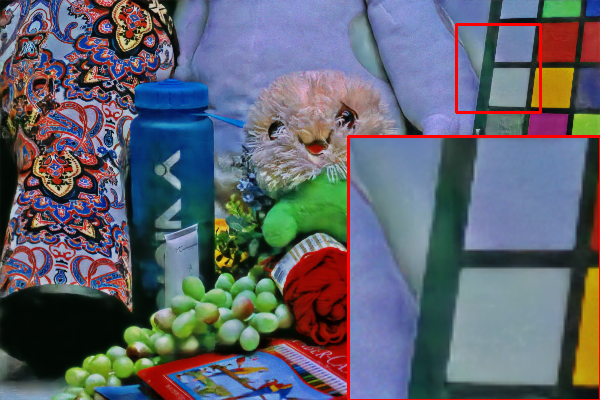}
}\hfill
\subcaptionbox{SNR-Net\label{fig:comparison_lol_snr}}{
  \includegraphics[width=0.15\linewidth]{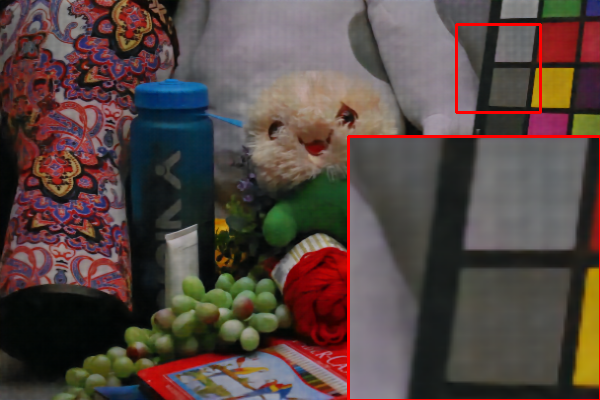}
}\hfill
\subcaptionbox{ZeroDCE\label{fig:comparison_lol_dce}}{
  \includegraphics[width=0.15\linewidth]{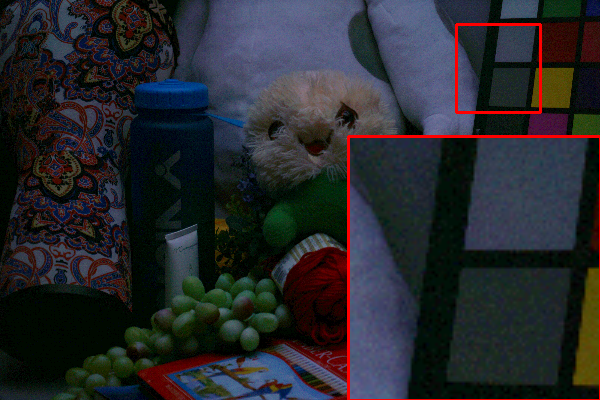}
}\hfill
\subcaptionbox{EnGAN\label{fig:comparison_lol_eng}}{
  \includegraphics[width=0.15\linewidth]{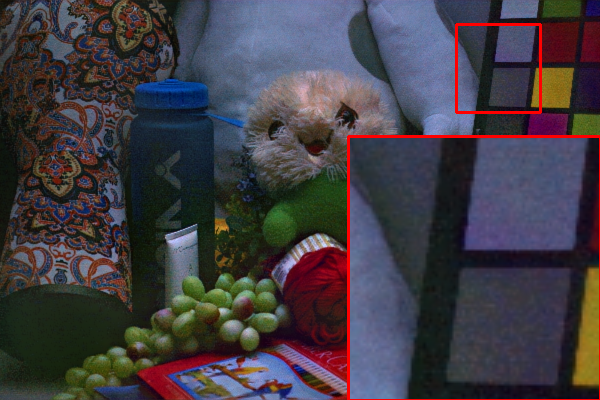}
}\hfill
\subcaptionbox{RUAS\label{fig:comparison_lol_ru}}{
  \includegraphics[width=0.15\linewidth]{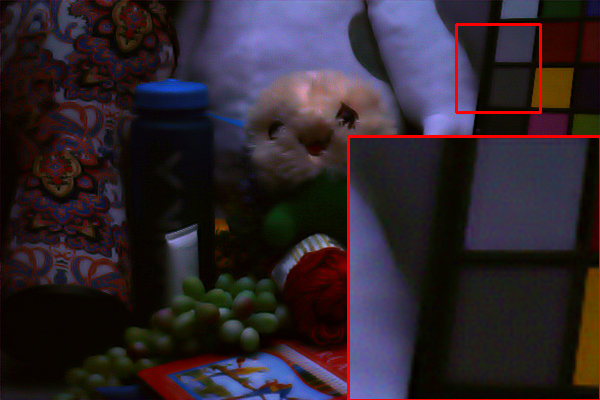}
}

\subcaptionbox{SCI\label{fig:comparison_lol_sci}}{
  \includegraphics[width=0.15\linewidth]{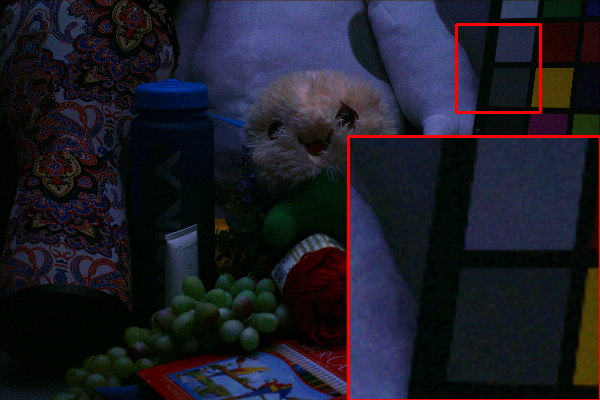}
}\hfill
\subcaptionbox{PairLIE\label{fig:comparison_lol_palie}}{
  \includegraphics[width=0.15\linewidth]{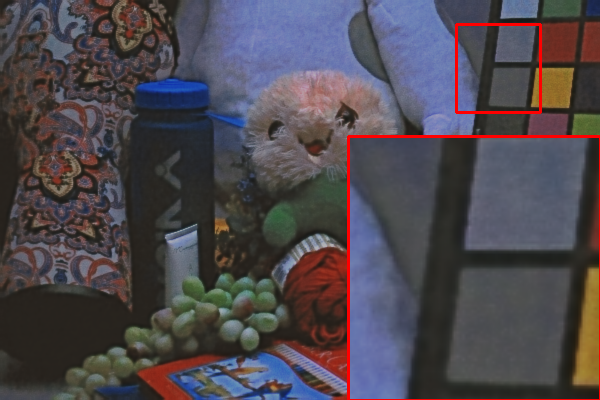}
}\hfill
\subcaptionbox{Ours\label{fig:comparison_lol_ours}}{
  \includegraphics[width=0.15\linewidth]{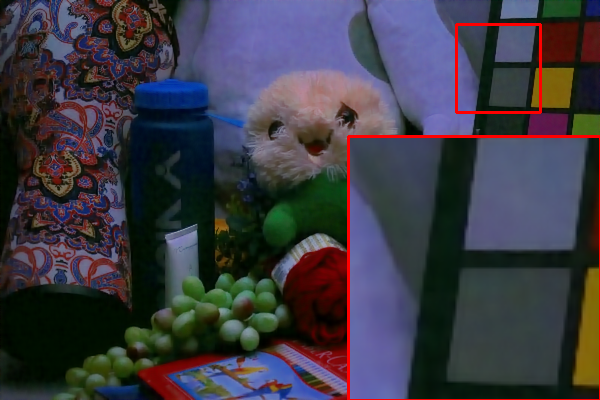}
}\hfill
\subcaptionbox{Ours-Small\label{fig:comparison_lol_ours_small}}{
  \includegraphics[width=0.15\linewidth]{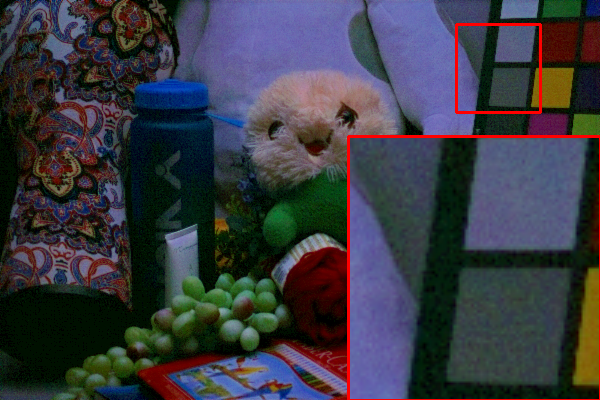}
}\hfill
\subcaptionbox{Ours-Tiny\label{fig:comparison_lol_ours_tiny}}{
  \includegraphics[width=0.15\linewidth]{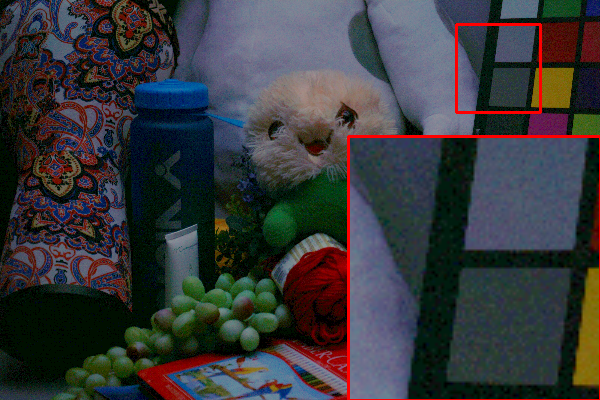}
}\hfill
\subcaptionbox{Ground Truth\label{fig:comparison_lol_gt}}{
  \includegraphics[width=0.15\linewidth]{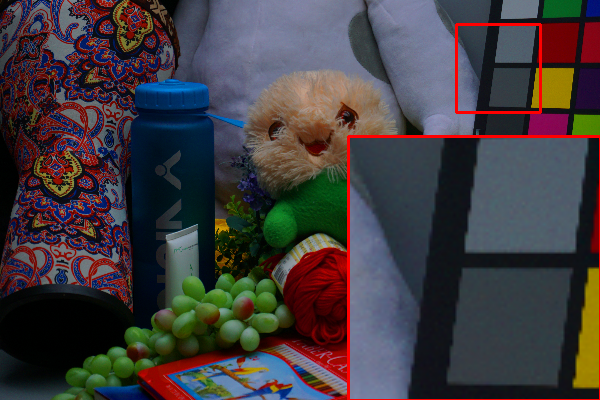}
}

\caption{Visual Comparison on LOL dataset.}
\label{fig:comparison_lol}
\end{figure*}

\begin{figure*}[tb]
\centering
\captionsetup{skip=2pt} 
\captionsetup[subfigure]{skip=2pt} 
\subcaptionbox{Input\label{fig:org_99}}{
  \includegraphics[width=0.15\linewidth]{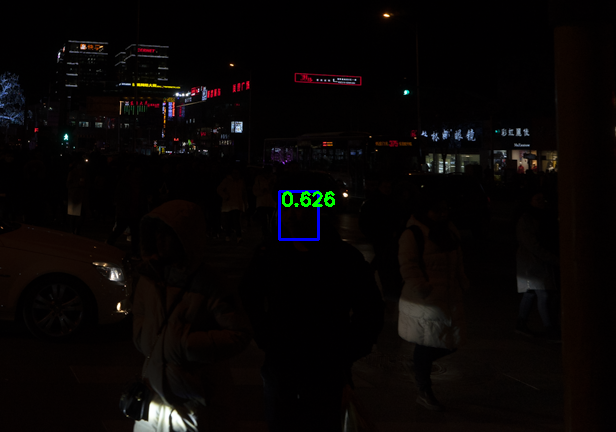}
}\hfil
\subcaptionbox{URN\label{fig:urn_99}}{
  \includegraphics[width=0.15\linewidth]{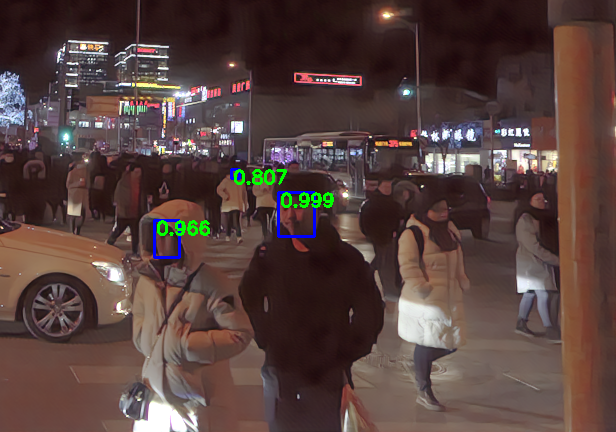}
}
\subcaptionbox{SNR-Net\label{fig:snr_99}}{
  \includegraphics[width=0.15\linewidth]{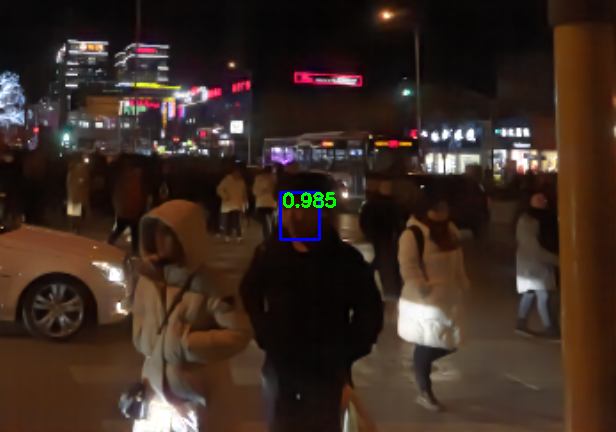}
}\hfil
\subcaptionbox{Retinexformer\label{fig:rf_99}}{
  \includegraphics[width=0.15\linewidth]{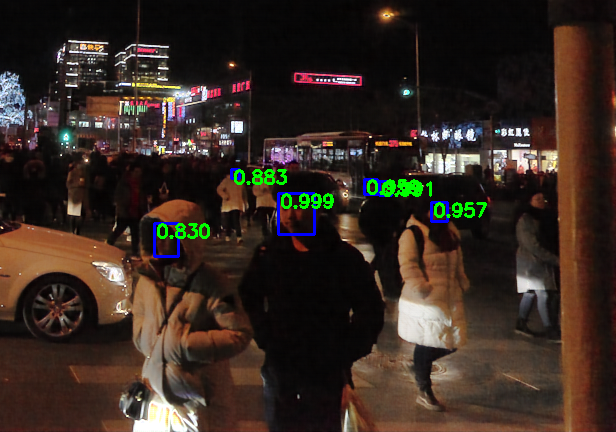}
}\hfil
\subcaptionbox{ZeroDCE\label{fig:dce_99}}{
  \includegraphics[width=0.15\linewidth]{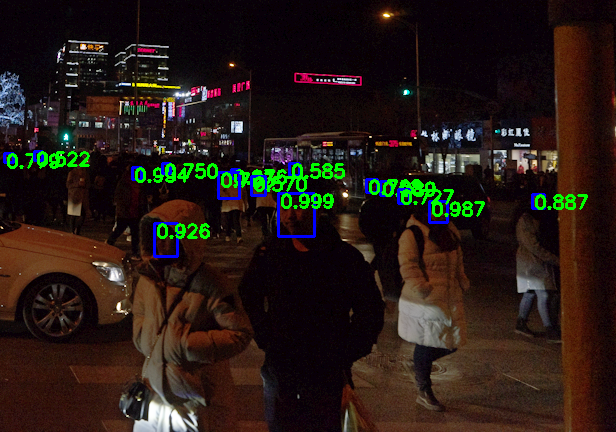}
}\hfil
\subcaptionbox{EnGAN\label{fig:eng_99}}{
  \includegraphics[width=0.15\linewidth]{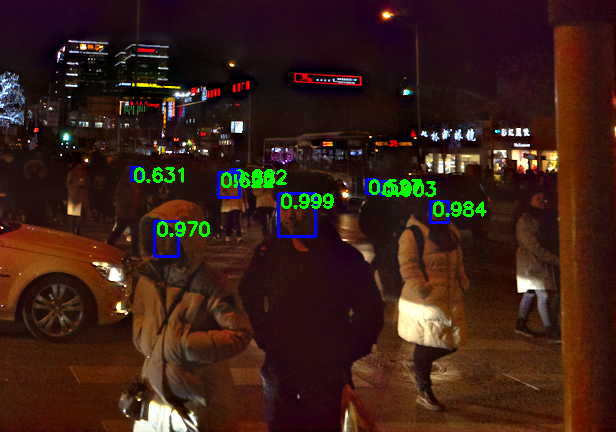}
}

\subcaptionbox{RUAS\label{fig:ru_99}}{
  \includegraphics[width=0.15\linewidth]{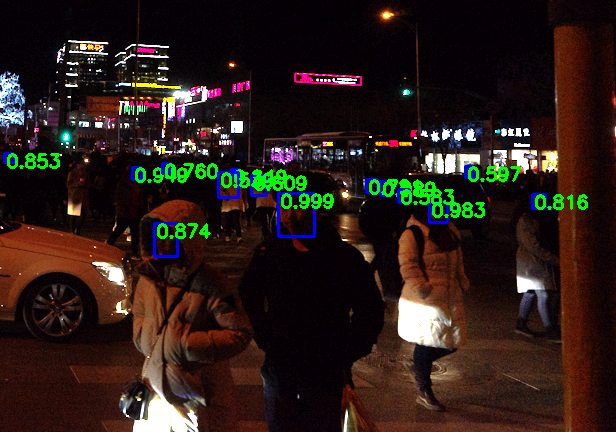}
}
\subcaptionbox{SCI\label{fig:sci_99}}{
  \includegraphics[width=0.15\linewidth]{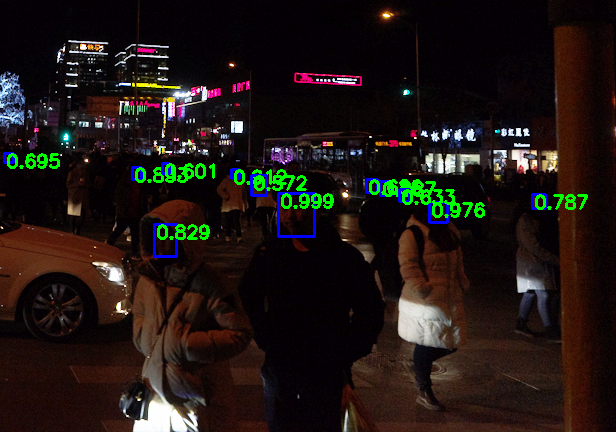}
}\hfil
\subcaptionbox{PairLIE\label{fig:pa_99}}{
  \includegraphics[width=0.15\linewidth]{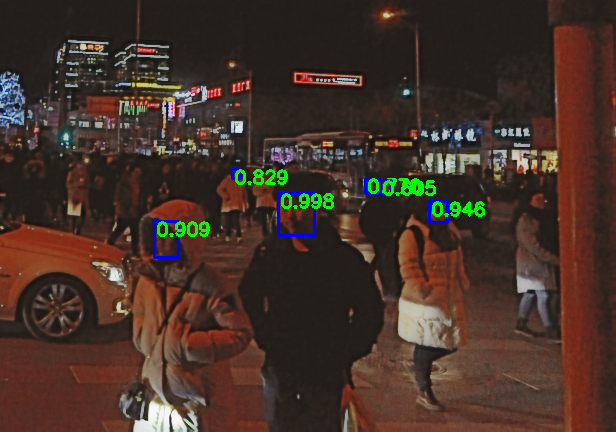}
}\hfil
\subcaptionbox{Ours-Small\label{fig:ours_s_d_99}}{
  \includegraphics[width=0.15\linewidth]{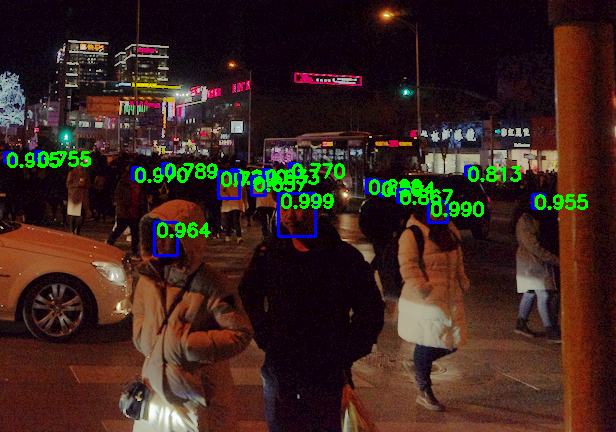}
}\hfil
\subcaptionbox{Ours-Tiny\label{fig:ours_d_99}}{
  \includegraphics[width=0.15\linewidth]{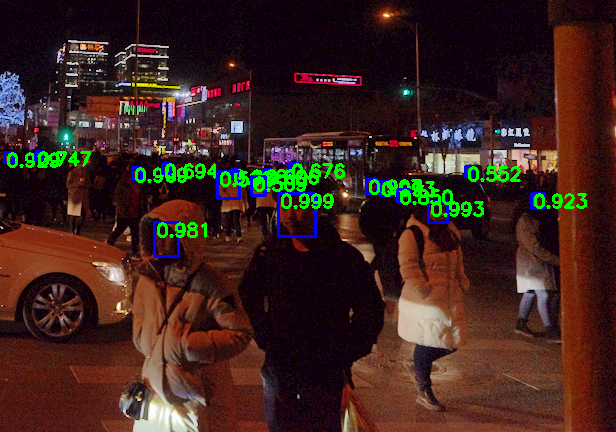}
}\hfil
\subcaptionbox{Ground Truth\label{fig:gt_99}}{
  \includegraphics[width=0.15\linewidth]{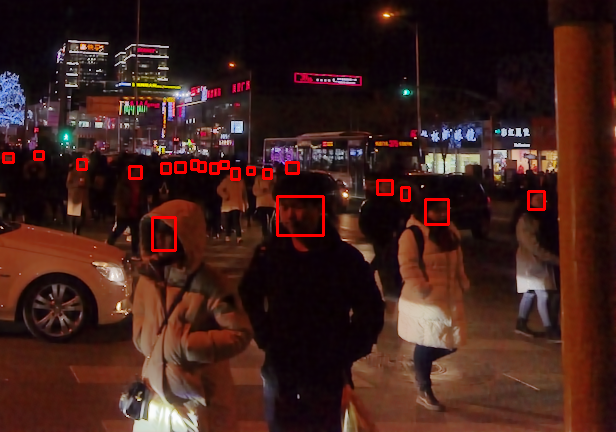}
}

\caption{
Visual comparison of face detection on the DarkFace dataset. The blue box shows the detected human face, while the red box marks the GT. The number represents the confidence score from Retinaface, with only scores above 0.50 displayed.
}
\label{fig:comparison_face_99}  
\end{figure*}

\subsection{Comparison with the State-of-the-art}

We conduct a comprehensive evaluation on the LOL-test~\cite{Chen2018Retinex} and SCIE-part2~\cite{cai2018learning} datasets. For SCIE-part2, we select the image from each of the first 100 subsets and resize them to $512 \times 512$. To assess the generalizability, our models are trained solely on the LOL dataset and directly tested on SCIE-part2 without any fine-tuning.
We employ four metrics to evaluate performance: PSNR and SSIM for structural fidelity, LPIPS for perceptual quality, and CIEDE2000 for color accuracy.
Furthermore, to demonstrate the practical utility of our method, we extend the comparison to the downstream face detection task on the DarkFace dataset~\cite{yuan2019ug}.
For fair comparison, the computational complexity (FLOPs) and inference speed (FPS) of all methods are measured on $3 \times 1200 \times 900$ RGB images. The FPS is evaluated on an NVIDIA A100 GPU.

\noindent\textbf{Image Quality Evaluation.} 
We compare our method against a wide range of SOTA algorithms, including supervised methods (KinD++~\cite{zhang2021beyond}, URN~\cite{wu2022uretinex}, SNR-Net~\cite{xu2022snr}, Retinexformer~\cite{Cai_2023_ICCV}) and unsupervised methods (RRDNet~\cite{zhu2020zero}, ZeroDCE~\cite{guo2020zero}, EnGAN~\cite{jiang2021enlightengan}, RUAS~\cite{liu2021retinex}, SCI~\cite{ma2022toward}, PairLIE~\cite{Fu_2023_CVPR}).
As reported in Tab.~\ref{tab:comparison}, our large-scale model (Ours) achieves the best performance among all unsupervised methods on the LOL-test dataset (PSNR 19.69 dB, SSIM 0.78 and LPIPS 0.18). 
Notably, our method demonstrates superior generalizability. While supervised methods like Retinexformer achieve high metrics on the LOL dataset, their performance drops drastically when transferred to the SCIE dataset (e.g., PSNR drops from 25.15 dB to 17.22 dB). In contrast, our approach maintains robust performance on the unseen SCIE dataset (21.02 dB), surpassing the cross-domain results of supervised counterparts.
In terms of efficiency, our lightweight versions, Ours-Small and Ours-Tiny, strike an excellent balance between performance and speed. 
For instance, Ours-Tiny (0.00034M parameters) achieves a PSNR of 17.65 dB on the LOL-test dataset and 19.85 dB on the SCIE-part2 dataset, outperforming (0.00035M parameters), which obtains 14.78 dB and 14.07 dB on LOL and SCIE, respectively. 
Moreover, Ours-Tiny maintains real-time inference speed at over 50 FPS.

\noindent\textbf{Visual Analysis.}
Visual comparisons on the LOL dataset are presented in Fig.~\ref{fig:comparison_lol}. 
Supervised methods often introduce artifacts (e.g., SNR-Net) or color shifts (e.g., KinD++), while unsupervised methods like ZeroDCE and EnGAN tend to amplify noise in dark regions.
In contrast, our method achieves more natural and balanced global brightness restoration compared with ZeroDCE, RUAS and SCI, effectively enhancing the visibility of dark areas without overexposure. 
As highlighted in the red box zoom-in views, our method also suppresses noise and preserves fine details compared to results of KinD++, SNR-Net, EnGAN, RUAS and PairLIE.

\noindent\textbf{Performance on Downstream Dark Face Detection.}
To validate our brightness and detail recovery, we evaluate downstream face detection using RetinaFace~\cite{deng2020retinaface} on the DarkFace dataset. As shown in Tab.~\ref{tab:comparison}, Ours-Small achieves the highest AP (0.666), significantly outperforming competitors. Notably, it surpasses supervised methods, which struggle to generalize due to the lack of paired ground truth for fine-tuning.
We also observe a trade-off between denoising and detection accuracy. While our larger model (Ours) offers better visual quality via stronger denoising, it yields lower AP (0.508) than lighter versions. This indicates that aggressive denoising smooths out critical high-frequency features, whereas texture-preserving models (Ours-Small/Tiny) favor detection. Fig.~\ref{fig:comparison_face_99} visually confirms that our method enables high-confidence detection in challenging scenarios.
\subsection{Ablation Study}

We conduct comprehensive ablation studies on the LOL-test dataset to validate the effectiveness of our method configurations and the contribution of each loss function component.

\noindent\textbf{Impact of Network Architecture.} 
As presented in Table~\ref{tab:abl-llae-hlas}, we investigate the trade-off between performance and model complexity by varying the number of channels and iterations for the LLAE and HLAS blocks.
The results indicate that reducing the channel dimension to 16 significantly limits the feature representation capability, leading to a performance drop. Conversely, increasing the channels to 64 or the LLAE iterations to 10 yields diminishing returns, offering negligible PSNR gains while increasing computational cost.
Furthermore, removing the HLAS module entirely results in a clear degradation (19.47 dB), confirming its necessity in our framework.

\noindent\textbf{Impact of Loss Components.}
Table~\ref{tab:loss_ab} analyzes the individual contribution of each loss term.
$L_{IL}$, $L_{RC}$, and $L_{CS}$ prove to be the most critical components. Removing $L_{IL}$, for instance, causes a catastrophic drop in PSNR to 7.91, demonstrating its fundamental role in global illumination recovery. Similarly, the absence of $L_{RC}$ and $L_{CS}$ leads to severe structural and chromatic distortions.
The remaining terms, including $L_{WB}$, SSIM loss, and gradient constraints, serve as essential refinements. While their individual quantitative impact is smaller (PSNR changes $< 1$ dB), they are indispensable for suppressing artifacts and preserving high-frequency details, ensuring the final high-fidelity output.

\section{Conclusion}
\label{sec:conclusion}

In this paper, we proposed Self-DACE++, a fast, lightweight, and unsupervised framework for low-light image enhancement. 
A key innovation of this work is the novel Adaptive Adjustment Curve, which enables flexible dynamic range adjustment while effectively suppressing artifacts and overexposure.
By integrating a physics-based Retinex loss with a randomized order training strategy, we successfully compress the network into an extremely lightweight iterative model without sacrificing performance.
Extensive experiments demonstrate that Self-DACE++ not only outperforms state-of-the-art methods in terms of visual quality and computational efficiency but also exhibits superior generalization capabilities on unseen datasets. 
Furthermore, our method serves as an effective pre-processing step, significantly boosting the performance of face detection in low-light scenarios.

\bibliographystyle{IEEEtran}
\bibliography{refs}

\end{document}